\documentclass[letterpaper, 10 pt, conference]{ieeeconf}  

\IEEEoverridecommandlockouts                          
\let\labelindent\relax

\usepackage{times}         
\usepackage{amsmath,amssymb,amsfonts}  
\usepackage{mathtools}
\usepackage{gensymb}
\usepackage{textcomp}
\usepackage{hyperref}      
\usepackage{lipsum}        
\usepackage[shortlabels]{enumitem}

\usepackage{graphicx}      
\usepackage[export]{adjustbox}
\usepackage{subcaption}    
\usepackage[font=small]{caption}  

\usepackage{array}
\usepackage{tabularx}
\usepackage{booktabs}      
\usepackage{multirow}
\usepackage{multicol}
\usepackage{tablefootnote}
\usepackage{makecell}
\usepackage{tabularray}

\usepackage[table, dvipsnames]{xcolor}

\usepackage{xfrac}
\usepackage{float} 
\usepackage{breqn}
\usepackage{algpseudocode}
\usepackage[linesnumbered,ruled,lined]{algorithm2e}
\usepackage{etoolbox}

\usepackage[style=ieee, sorting=none, citestyle=numeric-comp, backend=biber,
  maxnames=3, minnames=1
]{biblatex}
\let\labelindent\relax  

\hypersetup{
    colorlinks=true,
    citecolor=red,
}

\AtEveryCitekey{\bibhyperref{\color{red}}}

\makeatletter
\def\@makefnmark{\hbox{\@textsuperscript{\normalfont\textcolor{red}{\@thefnmark}}}}
\makeatother

\begin{document}

\title{\LARGE \bf
Diffusion-FS: Multimodal Free-Space Prediction \\ via Diffusion for Autonomous Driving 
}

\author{Keshav Gupta$^{1}$, Tejas S. Stanley$^{1}$, Pranjal Paul$^{1}$, Arun K. Singh$^{2}$ and K. Madhava Krishna$^{1}$
\thanks{
    $^{1}$Robotics Research Center, IIIT-Hyderabad, India
}%
\thanks{$^{2}$The University of Tartu, Estonia}%
}

\makeatletter
\let\@oldmaketitle\@maketitle
\renewcommand{\@maketitle}{
    \@oldmaketitle
    \centering
    \includegraphics[width=\textwidth]{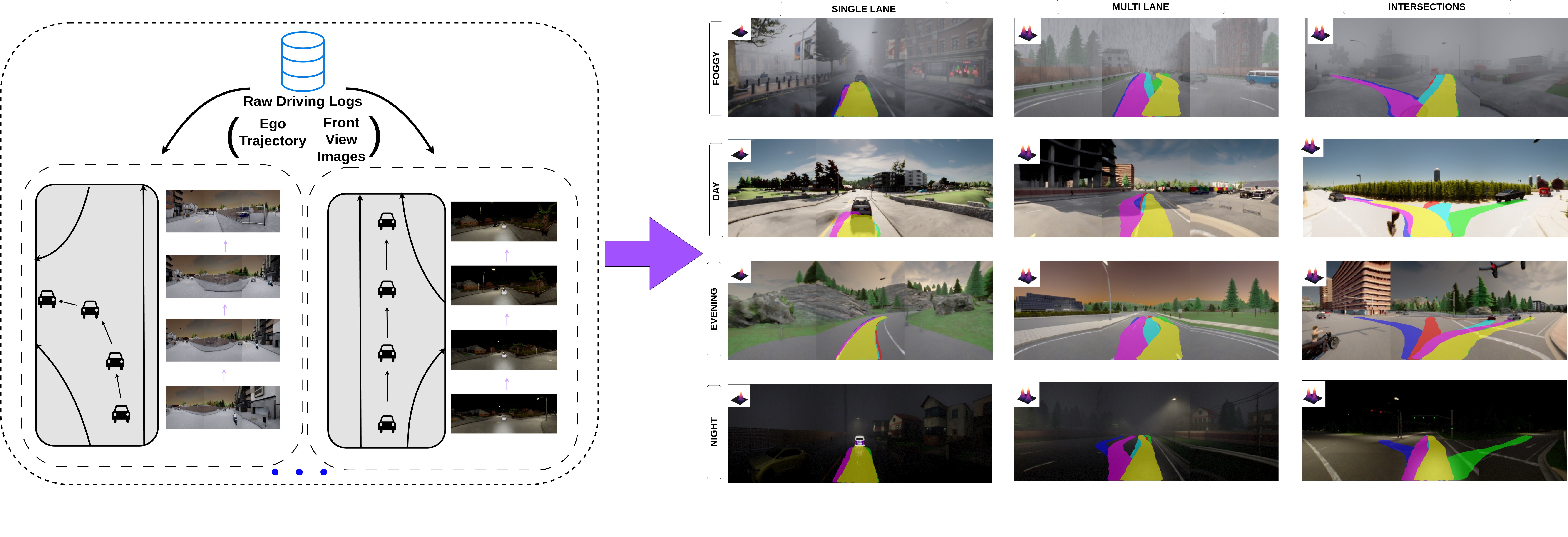}
    \captionof{figure}{\textbf{Left:} A dataset of raw driving logs containing image and ego trajectory pairs.
    Our self-supervised method processes such an unannotated dataset to generate free-space segments essential for autonomous driving.
    \textbf{Right:} Examples of multimodal free-space segments generated by our diffusion model on CARLA.
    At inference, our model denoises a fixed number of 6 noise samples into free-space segments.
    We showcase predictions across various weather conditions, times of day, road topologies, and obstacle layouts.}
    \label{fig:teaser}%
}
\makeatother

\maketitle

\begin{abstract}
Drivable Free-space prediction is a fundamental and crucial problem in autonomous driving. Recent works have addressed the problem by representing the entire non-obstacle road regions as the free-space. In contrast our aim is to estimate the driving corridors that are a navigable subset of the entire road region. Unfortunately, existing corridor estimation methods directly assume a BEV-centric representation, which is hard to obtain. In contrast, we frame drivable free-space corridor prediction as a pure image perception task, using only monocular camera input. However such a formulation poses several challenges as one doesn't have the corresponding data for such free-space corridor segments in the image.
Consequently, we develop a novel self-supervised approach for free-space sample generation by leveraging future ego trajectories and front-view camera images, making the process of visual corridor estimation dependent on the ego trajectory. We then employ a diffusion process to model the distribution of such segments in the image. However, the existing binary mask-based representation for a segment poses many limitations. Therefore, we introduce ContourDiff, a specialized diffusion-based architecture that denoises over contour points rather than relying on binary mask representations, enabling structured and interpretable free-space predictions. We evaluate our approach qualitatively and quantitatively on both nuScenes and CARLA, demonstrating its effectiveness in accurately predicting safe multimodal navigable corridors in the image.

Project Page - \href{https://diffusion-freespace.github.io/}{\textcolor{red}{https://diffusion-freespace.github.io/}}
\end{abstract}

\begin{figure*}[t]
\centering
\includegraphics[width=\textwidth]{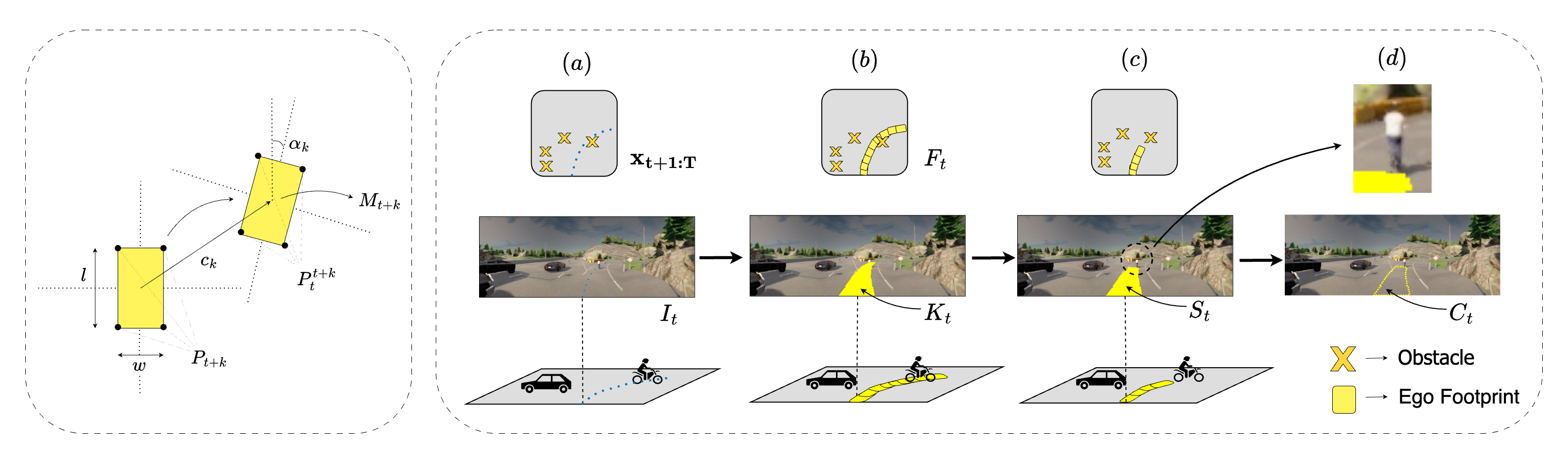}
\caption{\textbf{Free-space Contour Creation.} \textbf{Left : } We show the transformations between the ego vehicle's frame at time $t+k$ and the frame at time $t$. \textbf{Right}:  We show the process of creation of the free-space sample for an image. The top row presents the BEV map in the local frame of the ego vehicle at time $t$, while the middle and bottom rows show the corresponding frontal camera images and an alternative top-down view, illustrating the camera setup and projection process from a third-person perspective. (a) shows the future ego trajectory of the ego vehicle $x_{t+1:T}$. (b) shows the corresponding future footprint of the ego vehicle, $F_t$ and its projection $K_t$ in the image plane. (c) the future footprint $K_t$ is bounded up to the closest overlapping obstacle to obtain the free-space segment, $S_t$. (d) the corresponding free-space contour $C_t$ is obtained.}
\label{fig:freespace_sample_creation}
\end{figure*}

\section{Introduction} \label{introduction}

The autonomous navigation community is increasingly exploring vision-based approaches that aim to map the observations to actions either through direct perception \cite{shah2023vint, wu2022trajectoryguided, chitta2021neat} or via intermediate metric scene representation such as occupancy grid or BEV map \cite{philion2020lift, zeng2019end, hu2023planning}. 
The former often struggles with vehicle kinematic constraints, obstacle avoidance, and lane boundaries due to reliance on error-prone perception modules that map high-dimensional features to control inputs.
In contrast, humans while driving, identify \emph{free-space} 
\footnote{The free-space navigation in literature, is understood under two different categories: \textit{1. Drivable-Area Prediction} \cite{che2024twinlitenetplus, 9671392, hughes2019drivespace, zhao2017pyramid}, which represents the entire non-obstacle road region, broadly studied as lane segmentation and/or road segmentation task, and \textit{2. Driving-Corridor Estimation} \cite{kochdumper2024real, wursching2021sampling, manzinger2020using, orzechowski2019towards}, which discretises the drivable region into a reachable set where the vehicle can reach over time from its current ego position without collisions. Our interest lies in 2nd, posing it as a perception task. }
rather than enumerating and precisely localizing the obstacles. This free-space perception approach is evident in common driving scenarios. For instance, a driver merging onto a highway focuses on regions or \emph{``corridors``} between vehicles, rather than their exact position. For driving situations that need more informed decisions, the navigable space naturally diversifies into multiple potential paths or corridors, each representing a distinct mode of navigation. 
For instance, at a T-junction, the driver simultaneously perceives multiple valid navigable options - left turn, right turn, or proceeding straight. Each represents a distinct viable option where the ``correct" path depends on contextual factors such as destination intent or traffic flow that are inherently multimodal. Humans instinctively evaluate these multiple plausible paths before committing to one, relying primarily on a relative sense of depth perception regarding surrounding traffic agents, scouting out drivable spaces rather than making precise metric calculations.

Applying this intuition to autonomous driving, we explore the prediction of multimodal navigable regions directly from a monocular camera input. While the drivable area prediction task is not entirely novel \cite{che2024twinlitenetplus, 9671392, hughes2019drivespace, zhao2017pyramid, kochdumper2024real, wursching2021sampling, manzinger2020using, orzechowski2019towards}, studying them as vision-oriented task remains unexplored. 
Hence, we pose this problem as visual corridor prediction task which contrasts with existing work \cite{ kochdumper2024real, wursching2021sampling, manzinger2020using, orzechowski2019towards} that assumes prior knowledge of obstacle positions and adopt geometric or optimization-driven strategies to compute navigable regions. 

\begin{figure}[!t]
\centering
\includegraphics[width=0.4\textwidth]{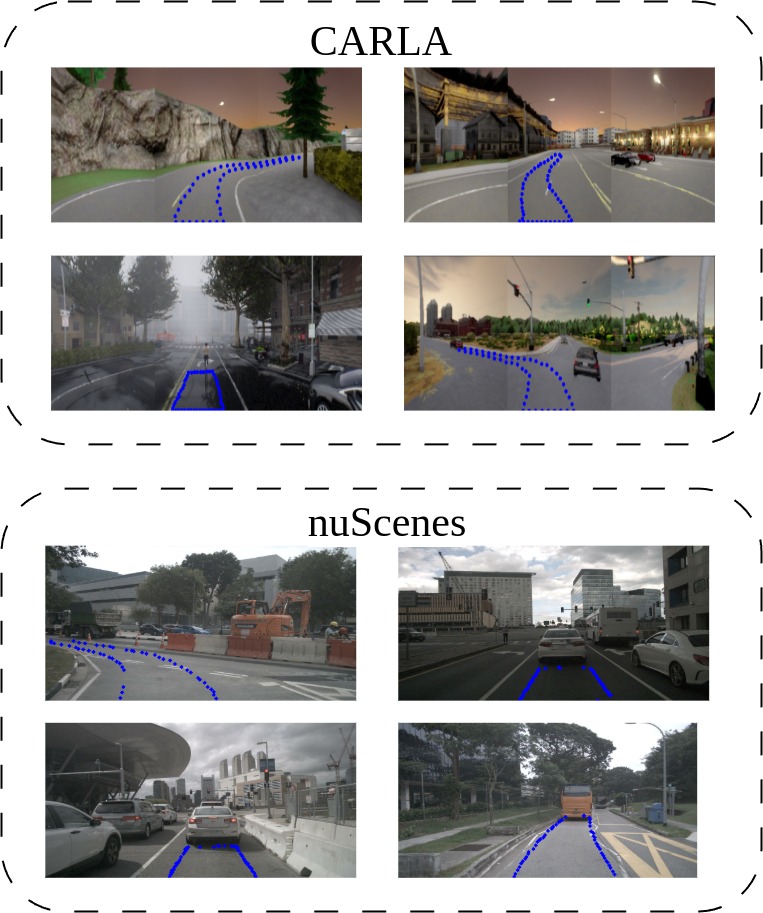}
\caption{\textbf{Training Samples :} We show different samples generated on CARLA and nuScenes, on applying the methodology described in Section \ref{methodology}}
\label{fig:data-sample}
\end{figure}

In this paper, we define navigable regions as pixel-level segments in terms of contour points that are a set of collision-free regions in the vicinity of the vehicle, as shown in Fig. \ref{fig:data-sample}.

Our contributions are the following:
\begin{LaTeXenumerate}
    \item We formulate the task of \textbf{visual corridor prediction} as an image perception task for the first time, contrary to prior works that assume the availability of a BEV centric representation. Our method predicts drivable paths directly from the ego vehicle's viewpoint, using front-view images to determine feasible corridors while avoiding obstacles and off-road areas.
    \item We propose a novel \textbf{self-supervised} approach for free-space sample generation from future ego trajectory and images, removing the dependence on dense annotated data.
    \item \textbf{ContourDiff} - We propose a novel diffusion architecture for denoising over contour points rather than a standard binary mask-based representation. Through the proposed architecture, we showcase for the first time, to the best of our knowledge, multimodal predictions of free space/traversable regions. These regions can be consumed by a downstream planner for navigation tasks. Further, by comparing both generative and non-generative segmentation models, we portray the superior performance of ContourDiff.
\end{LaTeXenumerate}

\section{Related Works} \label{related-works}



\subsection{Perception-Based Navigable Region Identification}
Perception-driven methods for identifying navigable regions in autonomous driving have been widely explored. 
Free-space segmentation approaches \cite{che2024twinlitenetplus, 9671392,, hughes2019drivespace, zhao2017pyramid} classify entire roads as free-space, losing the essence of true navigable regions. Their reliance on supervised learning with labeled datasets limits generalization to diverse road structures. Also, works like \cite{hosomi2024trimodal, paul2024lego} map linguistic commands to goal regions rather than segmenting navigable space. Overcoming these limitations, DiffusionFS adopts a diffusion-based approach to generate multimodal predictions of navigable corridors while maintaining semantic awareness of the surrounding environment.
By explicitly avoiding obstacles and off-road areas, it constructs feasible driving corridors from the ego vehicle’s perspective.

\subsection{Diffusion-Based Segmentation Approaches}
Recent advances in diffusion models have enabled their application in segmentation by leveraging generative processes to create or refine segmentation masks. Unlike standard segmentation approaches \cite{khanam2024yolov11, ren2015faster, redmon2016you}, which directly classify pixels, these methods utilize learned diffusion processes to generate structured masks or extract meaningful features from the denoising steps. SegDiff \cite{amit2021segdiff} models segmentation as a conditional generation task using Conditional Diffusion Probabilistic Models. However, due to the high dimensionality of segmentation masks, it converges slowly, as it denoises full-resolution masks, making it impractical for autonomous driving. To overcome this, ContourDiff operates directly on contour representations instead of full-resolution masks. This significantly reduces dimensionality and increases the convergence time, enabling efficient segmentation without compromising the fidelity of navigable regions. Other approaches \cite{Wu_2023_ICCV, baranchuk2022label} for segmentation via diffusion depend on large-scale image diffusion models \cite{rombach2022high}, which are highly computationally intensive and impractical for autonomous driving applications. Additionally, diffusion-based image segmentation has been explored in medical imaging \cite{wu2024medsegdiff, Rahman_2023_CVPR}, where models are based on the similar architecture as \cite{amit2021segdiff}, adapted to domain-specific challenges. 

\section{METHODOLOGY} \label{methodology}

\begin{figure}[!t]
\centering
\includegraphics[width=0.5\textwidth]{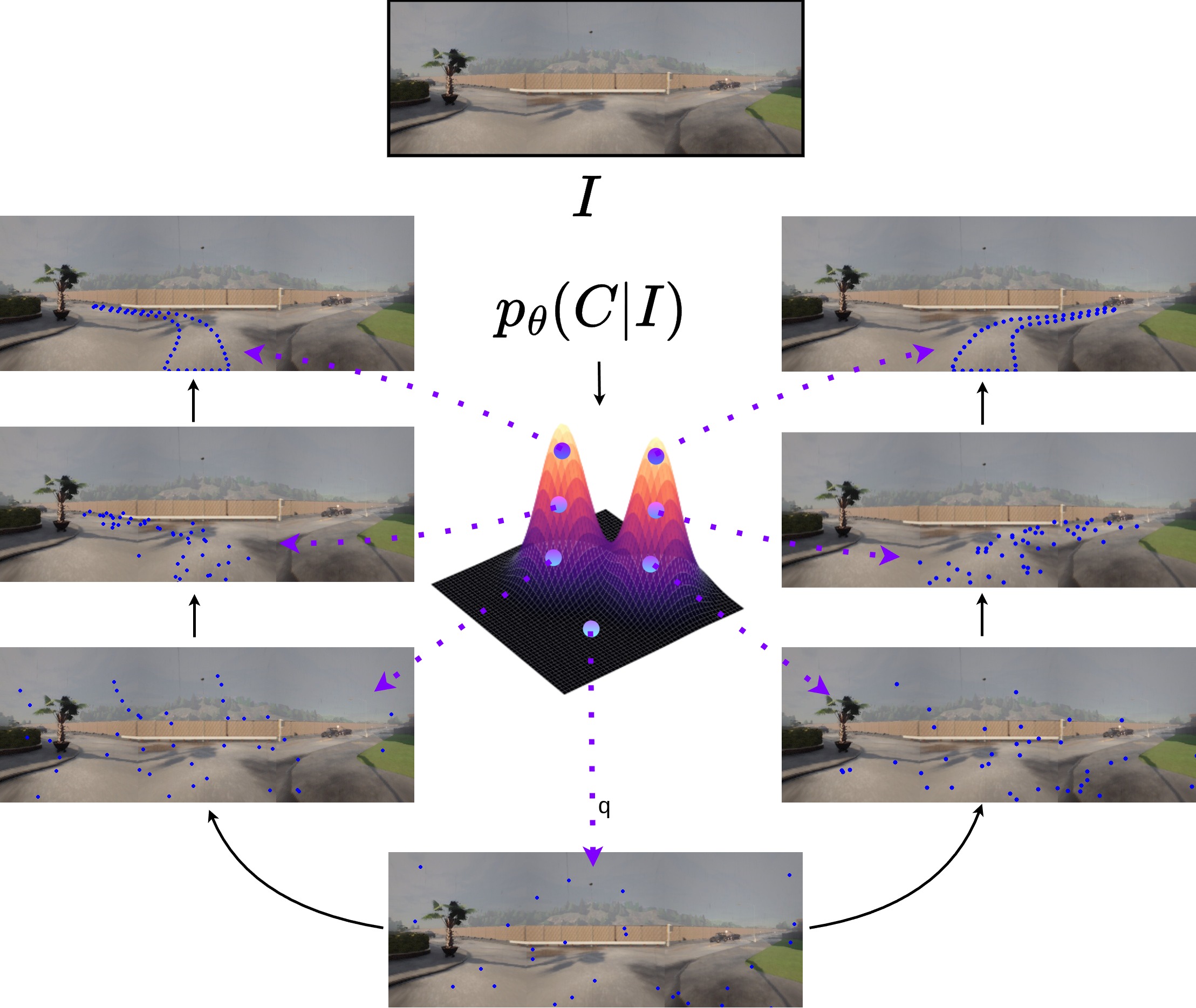}
\caption{\textbf{Conditional Probability Distribution of Free-space Contours given an image.}
We show an example of an intersection where the distribution of free-space contours is likely to be bimodal, as there is possibility of free-space at both the left and the right turn. The training data provides enough evidence to approximate this distribution, as in many driving logs covering a similar scenario, the ego vehicle must have traversed along both ways.}
\label{fig:diffusion_motivation}
\end{figure}

We propose a novel self-supervised method for directly detecting free-space in images without requiring any annotated data. Instead, our approach leverages raw driving logs of the ego vehicle, which are naturally abundant, extensive, and easily accessible from large-scale autonomous driving datasets or onboard vehicle sensors. Using a diverse set of pairs of images and corresponding free-spaces obtained through this method, we then train a conditional diffusion model over the free-space contours. During inference for an image, one can sample multiple free-space segments allowing for scalable, annotation-free, and adaptable free-space prediction.

\begin{figure*}[t]
\centering
\includegraphics[width=\textwidth]{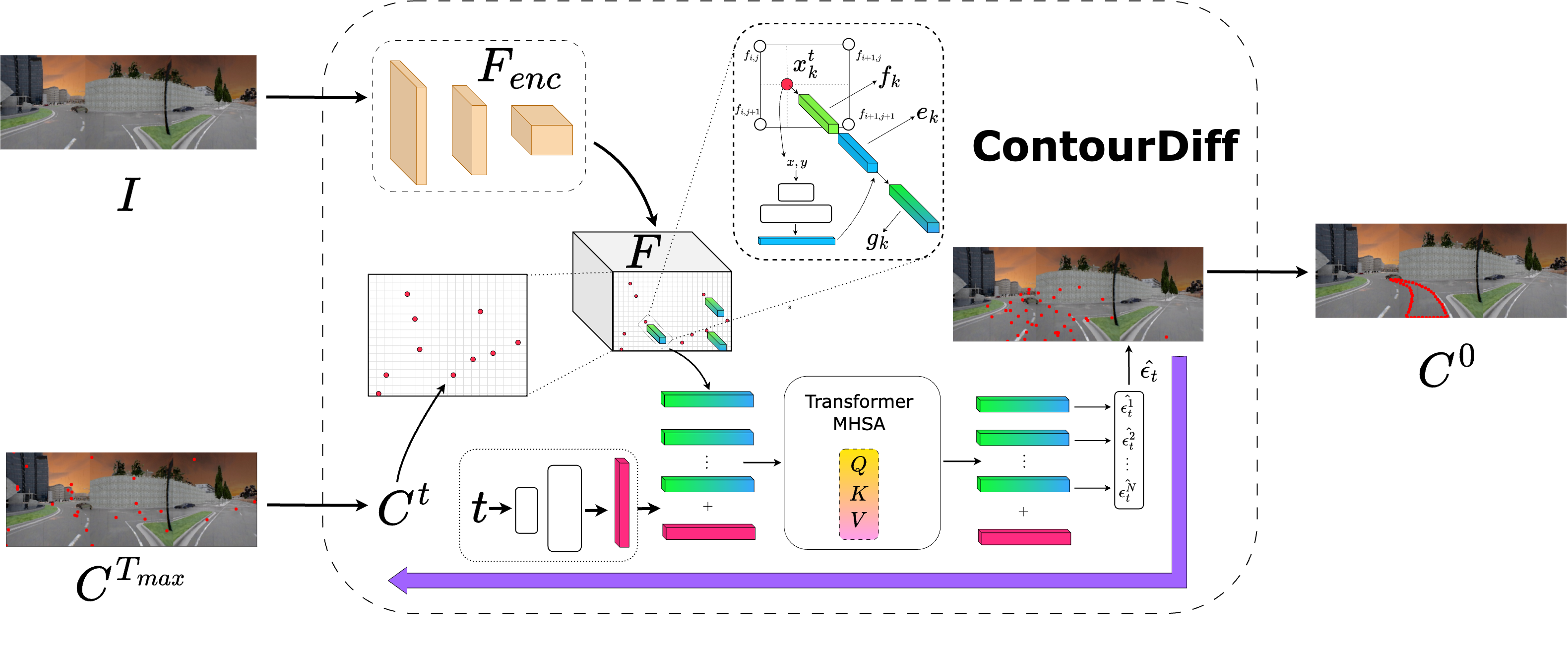}
\caption{\textbf{The architecture of the proposed ContourDiff.} The image $I$ and the initial noisy contour $C^{T_{max}}$ are passed as input to the model. Note that $C^t$ is visualized on top of image, and is not part of the image. The output of the model is the denoised contour $C^0$ which is obtained through running the reverse diffusion process.}
\label{fig:arch_diag}
\end{figure*}




\subsection{Self-Supervised Free-space Generation}

Our key observation is that the ego trajectory is inherently correlated with the free-space visible in the image, as we define free-space as the navigable portion of the road that aligns with human driving behavior. Since the ego vehicle always moves within a drivable region, its future positions provide a strong prior for free-space. We define one possible free-space segment by projecting the ego vehicle’s future footprints into the image. This segment does not entirely represent free-space, as the ego vehicle might have crossed regions in the future where some obstacle is present for the current timestep. This will yield an overlap of the free-space segment with the obstacle. We assume access to obstacle bounding boxes in the image plane, which can be efficiently obtained using existing object detection models, even for unannotated datasets using \cite{minderer2023scaling}. We limit the segment to the closest obstacle to guarantee free-space. As different driving episodes yield varying trajectories for similar scenes, our method captures multiple plausible free-space regions.



A driving log can be represented as a sequence of $\{(I_t, \mathbf{x}_t)\}_{t=1}^{T}$ pairs, where $I_t \in \mathbb{R}^{H \times W \times 3}$ is the image at the current timestep, $\mathbf{x}_t = \{x_t, y_t, \theta_t\} \in \mathbb{R}^3$ is the pose of the ego vehicle, and $T$ is the episode length. 
For a given timestep \( t \) in the driving log, our goal is to determine a possible free-space segment, which depends on the future trajectory of the ego vehicle, $\mathbf{x_{t+1:T}}$ = \( \{\mathbf{x}_{t+1}, \mathbf{x}_{t+2}, \dots, \mathbf{x}_{T}\} \).

For each future timestep $t+k$, we define the footprint of the ego vehicle, $M_{t+k}$ as the segment corresponding to the rectangular area of the ego vehicle in the local frame of the ego vehicle at timestep $t$, with the center of the rectangle as $$c_k = (x_{t+k} - x_t, y_{t+k} - y_t)$$

and the relative orientation as $$\alpha_k = \theta_{t+k} - \theta_{t}$$
Formally, $M_{t+k}$ is formed from the 4 points of the oriented bounding box corresponding to the footprint of the ego vehicle. Let $P_{t+k}$ define the set of corner points of the footprint in the frame of the ego vehicle at timestep ${t+k}$.

$$
P_{t+k} =
\begin{bmatrix}
    -\frac{w}{2} & \frac{w}{2} & \frac{w}{2} & -\frac{w}{2} \\
    -\frac{l}{2} & -\frac{l}{2} & \frac{l}{2} & \frac{l}{2}
\end{bmatrix}
$$
where $w, l$ are the width and length of the ego vehicle.
We next transform each of these points to the frame of the ego vehicle at timestep t denoted by $P_{t}^{t+k}$, by the transformation

$$P_{t}^{t+k} = R(\alpha_k) \cdot P_{t+k}  + c_k$$ 
where $R(\alpha_k)$ is the 2D rotation matrix corresponding to the angle $\alpha_k$. The above transformation is visualized in Fig. \ref{fig:freespace_sample_creation}. Let $p_1, p_2, p_3, p_4$ denote the transformed corner points i.e. columns of $P_{t}^{t+k}$. We get the transformed footprint mask $M_{t+k}$ as 
$$
M_{t+k}(u, v) = 
\begin{cases} 
1, & \begin{aligned} 
    \text{if } (u, v) \text{ is inside the rectangle formed} \\ 
    \text{by } \{p_1, p_2, p_3, p_4\}, 
    \end{aligned} \\
0, & \text{otherwise}.
\end{cases}
$$

The combined footprint mask, denoted as $F_t$ will be simply the bit-wise OR of these masks for all k. 
 $$F_t = \bigcup_{k=1}^{T-t} M_{t+k}$$

On getting $F_t$, we project this mask to the camera frame using known camera intrinsics and extrinsic parameters. For every point (u, v) in $F_t$, we get the projected point in the image frame using the camera intrinsics K, camera rotation matrix R and camera height h using 

$$
    \mathbf{p}' = K \cdot R \cdot 
    \begin{bmatrix} 
        u \\ 
        -h \\ 
        v
    \end{bmatrix}
$$
We denote this transformed mask as $K_t$. This mask represents the segmented path taken by the ego vehicle in future time-steps as viewed in the image plane in the current timestep, $t$. Given the set of bounding boxes of all the obstacles in the image plane, denoted as $O_t$, we limit $K_t$ to the nearest obstacle in $O_t$ by finding the closest obstacle overlapping with $K_t$, from which we get our desired free-space segment $S_t$. For our experiments, we deal with an equivalent conversion of this segment mask, which is the ordered set of contour points denoted as $C_t$. We get $C_t$ from $S_t$ using the implementation of \cite{suzuki1985topological}. Please refer to Fig. \ref{fig:freespace_sample_creation} for a visual illustration of the process.



\subsection{Diffusion Formulation}

Predicting the distribution of free-spaces in an image can be very challenging because of the complex multimodal nature of the task. To effectively model this, we use a diffusion model \cite{ho2020denoising} to approximate the true conditional distribution $q(C|I)$ of free-space contours $C$ given the image input $I$ through $p_\theta(C|I)$. Fig. \ref{fig:diffusion_motivation} presents an illustration of the approximated distribution expected through the diffusion process.



Starting from an initial noisy contour \( C^{T_{\max}} \sim \mathcal{N}(0, I) \), we iteratively denoise it to obtain a set of contours with decreasing noise levels, 
$
\{C^{T_{\max}}, C^{T_{\max}-1}, \dots, C^2, C^1, C^0\}
$. The denoising conversion from $C^t$ to $C^{t-1}$ follows the equation
\begin{equation}
C^{t-1} = \alpha(C^t - \gamma\epsilon_{\theta}(C^t, t) + N(0, \sigma^2I))
\end{equation}
where $\alpha$, $\gamma$ and $\sigma$ depend on the variance schedule of the diffusion process, and $\epsilon_\theta$ is the denoising model parameterized by $\theta$. 

During training, we have access to a dataset \( \mathcal{D} \) consisting of pairs of images and free-space contours, \( (I_i, C_i) \).  We sample the timestep $t$ uniformly from $(0, T_{max})$, and run the diffusion forward process for $t$ time-steps on the contour, $C$ from a randomly sampled pair $(I, C)$ from our dataset. The diffusion objective is 
\begin{equation}
    \min_{\theta} \mathbb{E}_{\substack{t \sim \mathcal{U}(0, T_{\max}) \\ \epsilon \sim \mathcal{N}(0, I) \\ (I, C) \sim \mathcal{D}}} 
    \left[ || \epsilon_{\theta}(I, C^t, t) - \epsilon||_2^2 \right]
\end{equation}

\subsection{ContourDiff - Denoising Contours}
Our proposed diffusion model architecture operates on contour points, offering a more interpretable alternative to diffusion over masks. Unlike mask-based diffusion, our approach ensures that at each timestep of the reverse process, we obtain a set of points directly on the image, providing a clear geometric interpretation of denoising. Such a property eliminates the need for thresholding to binarize the mask, as the point positions inherently define it. Additionally, representing a closed connected mask with points is both natural and efficient, requiring only $N\times2$ parameters compared to the $H\times W$ parameters needed for a latent mask representation. This structured representation acts as a prior for modeling connected closed segments, making it particularly well-suited for our task. This also improves the convergence during training and output quality. Refer to Fig. \ref{fig:arch_diag} for details about the architecture.
The input to the model is the set of noisy contour points at the forward process timestep t, 
$$C^t = \{x^t_1, x^t_2 ...x^t_N\} \in \mathbb{R}^{N\times2}$$
and the image $I$. For this, we pass the image through an image encoder $F_{enc}$ first to get the features corresponding to the image.
$$F = F_{enc}(I) \in \mathbb{R}^{H' \times W' \times D_f}$$
where $H', W'$ is the size of the downsampled feature map and $D_f$ is the dimension of each element of the feature map. \\
We then extract features at the noisy contour point locations using bilinear sampling, a method that interpolates features at fractional positions within the feature map. For each \( k \)-th point in the contour, \( x^t_k \), the bilinearly sampled feature is given by:
\[
f_k = \text{B}(F, x^t_k) \in \mathbb{R}^{D_f}
\]
where B is the sampling function that takes the feature map \( F \) and the sampling location \( x^t_k \) as inputs. The features lack any inherent position information, making it essential for the denoising model to be aware of each point’s location to accurately estimate the noise. To address this, we concatenate each point’s position with its sampled feature. Specifically, we first project the 2D position into a higher-dimensional positional embedding \( e_k \in \mathbb{R}^{D_e} \). Thus, for every point, we obtain the feature vector \( g_k \) as the concatenation of \( f_k \) and \( e_k \), forming a structured set of features corresponding to each contour point in \( C^t \).
\[G = \{g_1, g_2, \dots, g_N\} \in \mathbb{R}^{N \times D}\]
where  
\[
g_i =  
\begin{bmatrix} 
    f_i \\ 
    e_i 
\end{bmatrix} \in \mathbb{R}^D, \quad \forall i \in \{1, \dots, N\}, \quad D = D_f + D_e
\]
The timestep embedding $t_{emb}$ is taken as the standard sinusoidal positional embedding corresponding to timestep $t$. We then employ a series of transformer layers, incorporating multi-headed self-attention. We pass G along with the timestep embedding through these layers, enabling each point to capture dependencies with every other point as well as the timestep of the forward process. After the transformer layers, we use an MLP to map the \( D \)-dimensional embedding to the 2-dimensional observed noise $\epsilon_t$:
\[
H = \text{Transformer}(G, t_{\text{emb}}) \in \mathbb{R}^{(N+1) \times D}
\]
The predicted noise from the denoising model is given by:
\[
\hat{\epsilon_t} = \{ \text{MLP}(H_1), \dots, \text{MLP}(H_N) \} \in \mathbb{R}^{N \times 2}
\]

\section{Experiments and Results} \label{results}

Our experiments in this section are specifically designed to address the following questions:  
\begin{LaTeXenumerate}
    \item \textit{How does ContourDiff compare to prior works for segmentation via diffusion and other segmentation approaches?}
    \item \textit{What forms of conditioning or guidance can be added to the diffusion model to improve the quality of freespace segmentation?}
    \item \textit{How can we sample efficiently from the diffusion model to enhance multimodal outputs and have better control over the generated samples?}
\end{LaTeXenumerate}

\begin{figure*}[!t]
    \centering
    \includegraphics[width=1\textwidth]{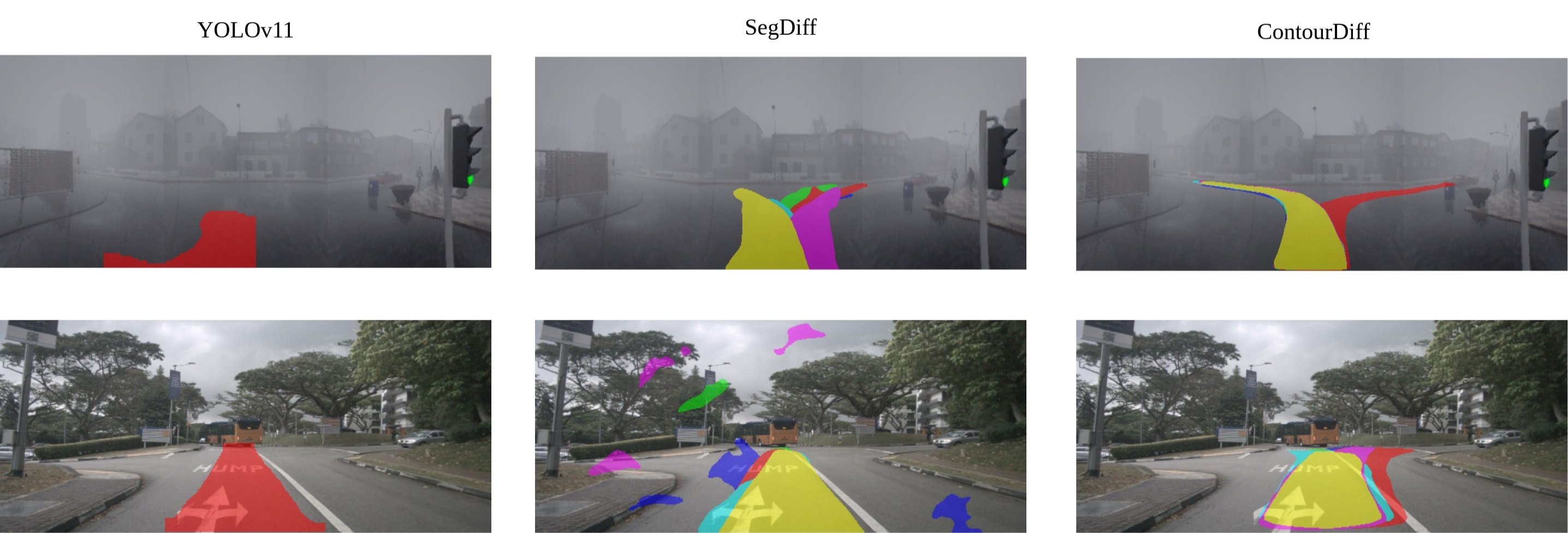}
    \caption{
    \textbf{Top Row: CARLA} – Comparison of YOLOv11, SegDiff, and our proposed ContourDiff at an intersection. The non-generative baseline YOLOv11 struggles to predict the free-space segment. We present six samples from both SegDiff and ContourDiff, demonstrating that ContourDiff, with its prior of points and more efficient parameterization, produces more refined and reasonable segments.
    \textbf{Bottom Row: nuScenes} – YOLOv11 outputs only a single free-space segment. SegDiff fails to generate connected samples as it has no prior for doing so unlike the contour representation. Hence it often predicts disconnected masks which undermines the task of free-space prediction. In contrast, ContourDiff significantly improves free-space segmentation, producing more accurate and diverse results.}
    \label{fig:comp1}
\end{figure*}

\subsection{Datasets}
We evaluate and benchmark our model on two datasets: one obtained through the CARLA simulator and nuScenes~\cite{caesar2020nuscenes}. Both datasets provide ego vehicle trajectory data, which we use to derive navigable free-space masks.
\begin{LaTeXenumerate}
    \item \textbf{CARLA:} We use the CARLA simulator to collect diverse training data via the method proposed by LAV~\cite{chen2022learningvehicles} data collection script. The dataset includes various driving scenarios such as straight roads, intersection turns, lane changes, and lane following. Our collected dataset comprises 82K frames from Towns 1–7, using three front-facing cameras with yaw angles of $-60^\circ$, $0^\circ$, and $60^\circ$. We split the dataset into 75K frames for training and 7K frames for evaluation.
    \item \textbf{nuScenes:} The nuScenes~\cite{caesar2020nuscenes} dataset provides real-world urban driving scenarios with ground-truth ego trajectories. We use its official split, consisting of 700 sequences for training and 150 for validation. We use only the front camera during the training process.
\end{LaTeXenumerate}

\vspace{-10pt}
\subsection{Implementation Details}
For CARLA, we stitch images from the three front-facing cameras, resulting in a final input size of $288 \times 768$. For nuScenes, the original image of size $900 \times 1600$ is resized to an input resolution of $256 \times 512$.
The model is trained with a learning rate of \(10^{-4}\) and a batch size of 64 across four NVIDIA RTX 3080 Ti GPUs (effective batch size: 256). The forward diffusion process follows a cosine beta schedule with $T_{max} = 50$ timesteps. The model predicts $N=50$ contour points, with features processed through six transformer blocks.
Our empirical observations indicate that all validation metrics converge up until 50 training epochs.
\begin{table}[b]
    \centering
    \caption{\textbf{Quantitative Results for Free-space Generation.}}
    \scriptsize 
    \renewcommand{\arraystretch}{1.1} 
    \begin{tabular}
    {@{} >{\centering\arraybackslash}p{1.5cm}
         >{\centering\arraybackslash}p{0.8cm} >{\centering\arraybackslash}p{0.8cm} >{\centering\arraybackslash}p{0.8cm}
         >{\centering\arraybackslash}p{0.8cm} >{\centering\arraybackslash}p{0.8cm} >{\centering\arraybackslash}p{1cm}@{}}

        \toprule
        Method & \multicolumn{3}{c}{CARLA} & \multicolumn{3}{c}{nuScenes} \\
        \cmidrule(lr){2-4} \cmidrule(lr){5-7}
        & \shortstack{IoU \\ ($\uparrow$)} & \shortstack{Obstacle\\Overlap \\ ($\downarrow$)} & \shortstack{Off-Road\\Overlap \\ ($\downarrow$)}
        & \shortstack{IoU \\ ($\uparrow$)} & \shortstack{Obstacle\\Overlap \\ ($\downarrow$)} & \shortstack{Off-Road\\Overlap \\ ($\downarrow$)} \\
        \midrule
        YOLOv11\cite{khanam2024yolov11} & 0.472 & 0.026 & 0.073 & 0.581 & \textbf{0.006} & 0.205 \\
        \midrule
        SegDiff\cite{amit2021segdiff} & 0.676 & 0.052 & \textbf{0.0032} & 0.628 & 0.061 & \textbf{0.022} \\
        ContourDiff & \textbf{0.7767} & \textbf{0.0200} & 0.0443 & \textbf{0.687} & 0.017 & 0.21 \\
        \bottomrule
    \end{tabular}
    \label{tab:fs_comp_table}
\end{table}

\subsection{Evaluation Metrics}
For free-space segmentation, we compute the mean \textbf{Intersection over Union (IoU)} between the predicted and ground truth free-space masks. To assess obstacle avoidance and safe navigation, we measure \textbf{Off-Road Overlap} which is the percentage of predicted free-space extending beyond the valid driving area and \textbf{Obstacle Overlap} which is the percentage of predicted free-space intersecting with detected obstacles.  

For CARLA, to analyze multimodality, we introduce an additional metric called \textit{Directional Deviation (DD)} to quantify variations in predicted samples. Specifically, we extract the centerline from the free-space mask by sampling the contour generated by the diffusion model. We then compute the angle of the line segment connecting the first and last points of the centerline. This process is repeated for six samples per image, and we calculate the mean and variance of these angles. Finally, we compute the average variance and average mean across the entire dataset to measure the diversity of the generated outputs. Additionally, we evaluate the Mean\textit{ Extent} of the angles, defined as the mean of the difference between the maximum and minimum angles among the six samples across the entire dataset, providing further insight into the spread of predictions.
\subsection{Comparing ContourDiff with other Segmentation Approaches}
As mentioned before, since our goal is to predict front-view safe navigable contours using diffusion, we found no prior works that define freespace as corridors in the image frame. To the best of our knowledge, this problem remains unexplored in existing research. Therefore, we establish two baselines:
\begin{LaTeXenumerate}
    \item \textbf{Non Generative - YOLOv11:} We train a YOLOv11\cite{khanam2024yolov11} segmentation model to demonstrate the limitations of a non-generative approach for this task.   
    \item \textbf{Generative - SegDiff:} To compare against a generative segmentation approach, we adopt SegDiff~\cite{amit2021segdiff}, a well-known diffusion-based image segmentation model which denoises over a standard mask-based representation,  as our baseline.
\end{LaTeXenumerate}
Table~\ref{tab:fs_comp_table} presents a comparison of segmentation evaluation metrics between the baselines and our proposed ContourDiff. Since a non-generative supervised model can only predict a fixed set of masks deterministically given an input image, YOLOv11 struggles to capture the inherent variability in navigable contour prediction. The poor performance on CARLA and nuScenes further highlights the necessity of a generative model for this task. 
The generative baseline shows our model outperforming SegDiff in IoU and obstacle overlap, highlighting the benefits of contour-based predictions over segmentation masks. Notably, ContourDiff maintains low off-road overlap. In contrast, SegDiff often predicts empty (null) masks, lowering both validation IoU and obstacle overlap, resulting in a lower obstacle overlap than ContourDiff.
Qualitative results are shown in Fig. \ref{fig:comp1}.



    

\begin{table}[b]
    \centering
    \caption{\textbf{Conditioning Results for Free-space Generation (CARLA).}}
    \begin{tabular}{@{}lccc@{}}
        \toprule
        Method & \shortstack{IoU \\ ($\uparrow$)} & \shortstack{Obstacle\\Overlap \\ ($\downarrow$)} & \shortstack{Off-Road\\Overlap \\ ($\downarrow$)} \\
        \midrule
        ContourDiff & \textbf{0.7767} & 0.0200 & 0.0443 \\
        Obstacle Guidance + ContourDiff & 0.7682 & \textbf{0.0098} & \textbf{0.0344} \\
        Class Conditioned ContourDiff & 0.6866 & 0.0252 & 0.0549 \\
        Noise Template + ContourDiff & 0.7718 & 0.0245 & 0.0501 \\
        Obstacle Guidance + Class Conditioned & 0.6851 & 0.0120 & 0.0424 \\
        \bottomrule
    \end{tabular}
    \label{tab:tr_carla_table}
\end{table}

\begin{table*}[htbp]
    \centering
    \caption{\textbf{Multimodality evaluation of Free-space Generation Across Road Scenarios (CARLA).}}
    \resizebox{\textwidth}{!}{%
    \begin{tabular}{@{}lcccccccccccccccc@{}}  
        \toprule
        \multirow{3}{*}{Method} & \multicolumn{4}{c}{NoLane} & \multicolumn{4}{c}{SingleLane} & \multicolumn{4}{c}{MultiLane} & \multicolumn{4}{c}{Intersection} \\  
        \cmidrule(lr){2-5} \cmidrule(lr){6-9} \cmidrule(lr){10-13} \cmidrule(lr){14-17}
        & IoU $\uparrow$ & \multicolumn{3}{c}{DD} & IoU $\uparrow$ & \multicolumn{3}{c}{DD} & IoU $\uparrow$ & \multicolumn{3}{c}{DD} & IoU $\uparrow$ & \multicolumn{3}{c}{DD} \\  
        \cmidrule(lr){3-5} \cmidrule(lr){7-9} \cmidrule(lr){11-13} \cmidrule(lr){15-17}
        &  & Mean  & Stddev  & Extent  &  & Mean  & Stddev & Extent &  & Mean & Stddev & Extent  &  & Mean & Stddev & Extent \\  
        \midrule
        Base Model  & 0.8158 & 95.56 & 3.03 & 8.57 & 0.7633 & 92.96 & 3.68 & 10.54 & 0.7762 & 94.52 & 3.04 & 8.83 & 0.7447 & 85.95 & 4.11 & 11.53 \\  
        Noise Template & 0.8100 & 96.67 & 4.22 & 12.33 & 0.7575 & 92.67 & 5.06 & 14.65 & 0.7787 & 94.16 & 3.48 & 10.20 & 0.7166 & 85.57 & 9.07 & 25.81 \\  
        Class Conditioning & 0.7326 & 96.45 & 8.12 & 23.15 & 0.6758 & 93.62 & 7.60 & 21.92 & 0.6854 & 94.16 & 5.60 & 16.41 & 0.6430 & 87.00 & 13.21 & 36.87 \\  
        Obstacle Guidance & 0.8155 & 95.57 & 3.10 & 8.81 & 0.7549 & 93.08 & 3.77 & 10.80 & 0.7630 & 94.47 & 3.17 & 9.19 & 0.7403 & 85.97 & 4.18 & 11.74 \\  
        Obstacle Guidance + Class Conditioned & 0.7350 & 96.67 & 8.02 & 23.03 & 0.6742 & 93.68 & 7.54 & 21.78 & 0.6823 & 94.19 & 5.65 & 16.55 & 0.6424 & 87.31 & 13.03 & 36.38 \\  
        \bottomrule
    \end{tabular}%
    }
    \label{tab:tr_carla_table_dd}
\end{table*}


\begin{figure}[!t]
\centering
\includegraphics[width=0.5\textwidth]{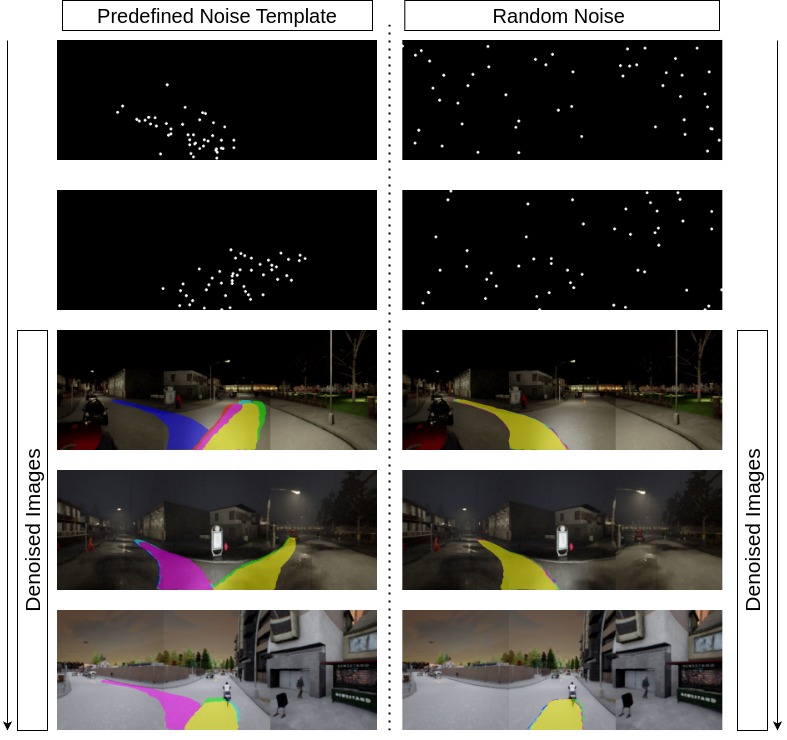}
\caption{\textbf{Effect of denoising from a set of predefined noise templates vs random noise template in the base model. Left:} Multimodality tends to increase as different modes are explored with different noise template initializations. \textbf{Right:} With random noise as initialization, the outputs tend to converge to a fixed sample.}
\label{fig:noise_template}
\end{figure}


\subsection{Enhancing Free-space Segmentation with Conditioning and Guidance}

\subsubsection{Class Conditioning - High Level Command Conditioning}
We examine the effect of conditioning on a class token representing broader driving behaviors, such as lane changes or turns, on the generation of multimodal predictions. For every frame, we have a label corresponding to one of the 6 high level commands. Possible high-level commands include turn-left, turn-right, go-straight, follow-lane, change-lane-to-left, change-lane-to-right. We one-hot encode the high-level command and project it to match the feature dimension of the transformer tokens. During training, we add this projected encoding to the set of input tokens. During inference, we sample a free-space segment for each of the six high-level commands, enabling diverse multimodal free-space predictions. 



As shown in Table \ref{tab:tr_carla_table}, the class conditioned model generates contours that align well with expected driving behavior. However, since the ground truth represents only a single specific behavior per scenario, the model’s diverse predictions-capturing multiple plausible behaviors are evaluated against a single reference. This inherently leads to a lower validation IoU compared to the base model.

However, this decrease in validation IoU does not indicate poor performance. Instead, it reflects the fact that the model is capable of generating multiple valid free-space predictions rather than being restricted to a single deterministic output. This ability to model multimodality is crucial in complex driving scenarios. 

Table \ref{tab:tr_carla_table_dd} further highlights the impact of class conditioning on both validation IoU and diversity. While the validation IoU is lower due to multimodal outputs, the diversity of free-space predictions is significantly higher across all cases compared to other methods. This demonstrates that class conditioning enables the model to produce a broader range of plausible free-space predictions, making it more adaptable to varying road scenarios in CARLA.
\subsubsection{Obstacle Guidance}
We evaluate the role of obstacle masks in guiding free-space prediction. The diffusion model is encouraged to avoid predicting contours inside obstacle regions by applying a correction gradient to points that fall within obstacles. The correction gradient points outside the obstacle and forces the contour points to move outside the mask.

As demonstrated in Table~\ref{tab:tr_carla_table}, obstacle guidance slightly reduces overlap with obstacles, ensuring that the predicted free-space aligns more closely with drivable regions.
\subsection{Efficient Sampling for Enhanced Multimodal Generation}
Motivated by the image editing technique in image diffusion models \cite{meng2022sdedit}, where noise is added iteratively for some timesteps to an input image and then denoised through the diffusion model to produce an edited version, we investigate the impact of spatially varying noise patterns on generating well-defined multimodal results. 

Fig.~\ref{fig:noise_template} right presents samples generated starting from random noise  alongside their corresponding starting noise. We observe that denoising from random noise often leads to convergence to fixed samples instead of exhibiting true multimodal behavior. This phenomenon is evident in Table~\ref{tab:tr_carla_table_dd}, where for the base model, the average extent at intersections, for example, is around 11.5 degrees.

To address this, we introduce structured initializations, allowing the model to reach local optima more effectively. As shown in Fig.~\ref{fig:noise_template}, we generate predefined noise templates by averaging \(K\) ground truth contours following a specific high-level command, \(K\) being a hyperparameter, and then applying the forward diffusion process for \(t\) timesteps. During denoising, we initialize from these templates and start the reverse denoising process from the same \(t\) timestep. The hyperparameter \(t\) is set to 10 in our experiments.

We find that initializing from six distinct noise templates, corresponding to six different high-level commands, significantly improves the multimodal behavior of the diffusion model while maintaining the other metrics, as demonstrated in Table~\ref{tab:tr_carla_table} and Table~\ref{tab:tr_carla_table_dd}, where both the extent and the variance are higher than that of the base model.

\section{Conclusion and Future Work} \label{conclusion}

In this paper, we present a self-supervised method for predicting visual corridors using diffusion models. Unlike previous approaches that rely on known BEV and/or obstacle locations, we treat the task as an image perception problem. We introduce a self-supervised strategy for generating free-space samples by utilizing future ego trajectories and images. Additionally, we create a contour-based diffusion architecture that focuses on denoising contour points rather than employing a binary mask, resulting in outputs that are more structured. We explore conditioning, guidance, and sampling techniques to enhance multimodality and control over the generated samples. Our results show ContourDiff's effectiveness in generating diverse, accurate free-space predictions, paving the way for future research in generative methods for autonomous navigation. We plan to validate and extend our approach to more challenging and diverse urban driving datasets.
\printbibliography
\end{document}